# Extracting Knowledge from an Arabic-English Machine-Readable Dictionary Using Information Extraction

Diaa Mohamed Fayed, Aly Aly Fahmy
Department of Computer Science
Faculty of Computers and Information, Cairo University
Giza, Egypt
diaa.fayed@outlook.com, a.fahmy@fci-cu.edu.eg

Mohsen Abdelrazek Rashwan
Department of EECE
Faculty of Engineering, Cairo University
Giza, Egypt
mrashwan@ieee.org

Wafaa Kamel Fayed
Department of Arabic Language and Literatures
Faculty of Arts, Cairo University
Giza, Egypt
wafkamel@link.net

***Abstract*— Natural language processing (NLP) applications need large and rich amount of linguistic knowledge. Furthermore, electronic language sources such as dictionaries, encyclopedia, and corpora became available. So, automatic methods are emerged to extract lexical information from those sources to overcome the knowledge acquisition bottleneck. We presented a method to automatically extract lexical information from a machine-readable version of the Arabic-English Al-Mawrid dictionary. We used n-gram analysis and key-word-in-context (KWIC) analysis to discover lexical patterns that manifest morphologic, syntactic, or semantic information. Then, we used hand-crafted rule-based information extraction to extract that information. Furthermore, we used punctuation marks and some heuristics to extract a set of synonyms in a subentry. This study registered high precision for all types of information, high recall for synonyms, and low recall for the other information. The study also showed that the Al-Mawrid has significant amount of derivations (morphologic information) and synonyms, domain labels, and hyponym/hypernym relations (semantic information).**

***Keywords—Arabic; English; Al-Mawrid; Machine Readable Dictionary; Computational Lexicography; Information Extraction; Rule-based; Hand-crafted;Hand-coded; Regular expressions***

## I. INTRODUCTION

Machine-readable dictionaries (MRDs) are electronic versions of ordinary printed dictionaries. Publishers use MRDs to produce printed dictionaries. Fontenelle [1] differentiated between *computerized dictionaries* and *machine-readable dictionaries*. The computerized dictionaries are organized logically, so they are easily processed or transformed into lexical databases (LDBs) where each piece of information (definitions, parts-of-speech, grammar codes, etc.) is partly formalized and identified. On the other hand, the machine-readable dictionaries are flat text files with codes for driving the typesetting process. The computerized dictionaries have many applications [2] such as machine translation, natural language comprehension, pronunciation guides for speech recognition, etc.

MRDs research became inactive field because the researchers shifted to the corpora [3, 4] and the web [5-7] as better sources of lexical information. However, this type of researches is still needed for poor-resource languages and for languages that have rich heritage of dictionaries that need to be computerized in order to be exploited in NLP applications. The Arabic language has both little NLP resources and much heritage of printed, not computerized dictionaries.

Building practical NLP applications need large and rich amount of lexical knowledge about words. A lexicon contains that information is the crucial component of any knowledge-based NLP systems. Unfortunately, constructing such lexicons manually is time and money consumption task. The need for large amounts of knowledge for practical NLP systems is called bottleneck knowledge problem in the Artificial Intelligence (AI). To solve the bottleneck knowledge problem, automatic extraction methods of lexical information from machine-readable sources such dictionaries, encyclopedia, and corpora was emerged.

Automatic acquisition of lexical information from MRDs has two main approaches: *text processing* and *syntactic parsing*. Montemagni and Vanderwende [8] described the text processing as “pattern matching at the string level (lexical patterns)” and described syntactic parsing as “pattern matching at the structural analysis level (structural patterns)”. Each approach has advantages and disadvantages. Refer to [9-12] for more comparisons and discussions.

Text processing methods do not use any of NLP tools such as morphological analyzer and syntactic parser to analyze the text. Text processing methods usually use pattern matching technique or n-gram analysis technique to discover lexical patterns that manifest morphologic, syntactic or semantic attributes or relations. Examples of studies that used text processing approach are [13-17]. On the other hand, syntactic parsing methods use either general-purpose (or broad-coverage) syntactic parser or special-purpose (or specially constructed) syntactic parsers. In these methods, the definitions text of a dictionary is parsed syntactically. Then, the output trees of the parser are analyzed to discover any syntactic or semantic relations. Examples of specially constructed parsers method are [2] [18] [19, 20] [9, 21]. Examples of the broad coverage text parsers method are [11] [22] [23] [24, 25] [26].

In this study, we assume that the Al-Mawrid definitions contain significant amount of linguistic information that are useful for NLP applications and that can be extracted automatically. Apart from the main approaches to automatically extract lexical information from MRDs, we adopted to apply the information extraction (IE) as a novel method in the MRDs research for some reasons: (a) we have not a general-purpose syntactic parser for the Arabic language; (b) building special-purpose syntactic parser for the Al-Mawrid is very expensive and is not the ideal option because the rich information variability of the Al-Mawrid definitions; and (c) the IE is more suitable than other methods to extract all types of randomly scattered information in the Al-Mawrid definitions.

The IE has two approaches: machine learning approach and rule-based approach. Although the rule-based information extraction approach is appeared to some researchers less elegant and less challenging than machine-learning approach, the study that was conducted by Chiticariu et al. [27] showed that the dominant IE systems, in the recent ten years, are rule-based IE systems. For more comparison between the two approaches, see [27].

We analyzed a sample of the Al-Mawrid using the n-gram analysis and key-word-in-context (concordance) analysis to discover lexical patterns that manifest morphologic, syntactic, or semantic information. Based on the discovered lexical patterns, we developed a set of IE rules to extract information from another sample of definition. We implemented the context-patterns of IE rules by regular expressions (REs). Furthermore, in another experiment, we extracted the synonymous words and phrases in each subentry.

The contributions of this paper are fourfold:

1. Applying the IE as novel, integrated method in the MRDs research.
2. Studding the structure and information of the Al-Mawrid computationally for the first time. Revealing the linguistic information in the Al-Mawrid definitions and categorizing them to morphologic, syntactic, and semantic information.
3. Showing some shortcomings in the Al-Mawrid from the respective NLP view.
4. Extracting linguistic information, structuring them, and then augmenting the Al-Mawrid database by new formalized linguistic knowledge.

## II. Related Works

Some lexical language resources are built manually for Arabic such as the Arabic WordNet [28] and [29]. A considerable number of researches are done in automatically extract linguistic information from Arabic, machine-readable data resources. Some works extracted semantic attributes or relations like [17, 30-33]. Others extracted multiword expressions (MWEs), multiword terms (MWTs), and noun compounds (NCs) like [34] [35] [36] [37]. Much works also are done on extracting Arabic named entities (NEs) from corpus such as [38] [39]. We note that most of Arabic works did not use dictionaries as sources of lexical information; the reasons for this may due to copyrights.

Diaa et al. [31] are the first team to study the Al-Mawrid definitions computationally. Diaa used Parsing Expression Grammars (PEGs) to capture the microstructure of the Al-Mawrid definitions and extract domain labels. Soha [17] extracted two semantic relations from the RDI machine-readable dictionary: synonym/hypernym and synonym relations. Soha used n-gram statistics technique to extract key phrase patterns or collocations. Then she used criteria to adopt phrases that suggest hyponym/hypernym semantic relations. Furthermore, she discovered some syntactic structures that may manifest synonyms or hyponym/hypernym semantic relations. Zamil and Al-Radaideh [30] proposed enhanced version of Hearst's algorithm to extract semantic relations from Arabic text. Lahbib et al. [33] used a vocalized corpus of hadith in three domains (Drinks, Purification, and Fasting) to extract semantic relations.

## III. The Al-Mawrid Structure

The Arabic-English Al-Mawrid version that is used in this study is general-purpose and medium-sized dictionary that has approximately 35000 distinct headwords and 60000 subentries. The Al-Mawrid was arranged alphabetically and articulately according to the first letters of the headwords. An entry of the Al-Mawrid starts by a bold headword. When a headword has more than one meaning or sense, each meaning occupies a subentry that is cited in a separated line. Subentries that express idioms, examples, restricted collocations, etc. are cited lately. A subentry consists of a mandatory Arabic section and an optional English section (translation). In addition, for distinguishing the headwords, each subentry text, except the first one, is shifted from the headword by two space characters. The Arabic section consists of three fields that we name *header*, *explanation* and *cross-reference*. The first field of an Arabic section is mandatory but the other two fields are optional. A colon separates the header from its explanation. A subentry may be expressed by *declaration*, *question*, or *exclamation phrases*.

The three fields of an Arabic section have the same structure: each field consists of one or more words or phrases that are separated by either an Arabic comma or the conjunction word "أو". These comma-separated phrases are mostly synonyms or near synonyms. The header has the headword of an entry if its subentry represents a sense of the headword. If a subentry does not represent a sense of the headword, it contains idioms, examples, restricted collocation, etc. The linguistic information is scattered randomly in the header or explanation fields. The English section has one or more translation equivalence groups (TEGs) that are separated by semicolons. Each TEG has one or more translation equivalence (TE) phrases that are separated by comma. The phrases of a TEG are synonyms. shows some entries of the Al-Mawrid.

|  |  |
|---|---|
| howl(ing), yelp(ing), yowl(ing) | عُوَاء (الكَلْبِ إلخ) |
| | عَوَائِد (مفردها عائِدَة) - راجع عائِدات |
| | عَوَائِق (مفردها عائِق) - راجع عائِق |
| | عَوَادِ (العَوَادِي) (مفردها عادِيَة) - راجع عادِيَة |
| lutist, lutanist | عَوَّاد: عازِفُ العُوْد |
| defect, fault, blemish | عَوَار، عُوَار: عَيْب، عِلَّة |
| anaphylaxis | عُوَار [طب] |
| swift, swallow, martin | عُوَّار (طائر) |
| average | عِوَار، عَوَارٍ، عَوَارِيَة [قانون بحري] |
| Herbivora, herbivores | عَوَاشِب: آكِلاتُ العُشْب |
| plankton | عَوَالِق: حُيَيْوِينَاتٌ أو نَبَاتَاتٌ طُحْلُبِيَّة |

Fig. 1. Entries examples of the Al-Mawrid.

The information in the Al-Mawrid may be explicit or implicit. Explicit information in dictionaries is the piece of information that is assigned to a code, symbol, or keyword (tag, label, etc.); the assigned information manifests the purpose or function of the information in the subentry. On the other hand, implicit information is not assigned to additional information, so the purpose can be inferred or deduced. Some analyses are needed for text definitions before extracting implicit information. The information that is to be extracted exists in the defining phrases of the explanation field or the header field. Fig. 1 shows some entries of the Al-Mawrid.

## IV. Theoretical Background

In this section, we give summaries about the MRDs research. Also, we give a summary of the information extraction methods and we explain the sublanguage analysis concept that is the theoretical foundation of our method to automatically extract linguistic knowledge from the Al-Mawrid dictionary by rule-based hand-crafted information extraction.

### A. Why is still MRDs Research important?

Although some happenings recently indicate the much decreasing of the research on MRDs, MRDs are still interesting for both practical and theoretical reasons. Some of practical reasons includes (a) MRDs can be used as the core constituent of a linguistic knowledge base and then the core is extended by combined information from other sources [40] such as corpora; (b) industrial NLP systems need larger and more sophisticated lexicons and the amount and types of the lexical resources required for them may differ from language to language [20]; (c) MRDs can provide NLP by good reusable lexical knowledge bases through reusability of the lexical and semantic information in the MRDs; and (d) MRDs research is still important for languages that have rare, electronic linguistic resources like Arabic [20].

### B. MRDs Research Difficulties

Although MRDs are structured lexical resources, some difficulties hinder extracting information from them. The following are some of difficulties [40]:

- *Physical Formats* (*Dictionary Format*): typesetter formats have some ambiguities and inconsistencies that hinder exploiting information directly from MRDs. Moreover, the diversity of conventions across published dictionaries is another problem.
- *Incoherence of Metatext* (*Defining Phrase*): there are inconsistencies in the structures of defining phrases that manifest semantic relations. These inconsistencies result from that the same semantic relation can be conveyed by an open set of structures in the same dictionary and across dictionaries.
- *The Bootstrapping Problem*: Acquiring robust lexical and semantic information from MRDs for NLP needs not only advanced syntactic parser, but also lexical information and word sense disambiguation. Then there is circularity. Some researchers begin to analyze MRD using simple hand-crafted NLP resources, and then bootstrap.

### C. Types of Knowledge in the MRDs

The information in MRDs is a subset of their printed copy since the MRDs are electronic versions of the ordinary, printed dictionaries. The MRDs contain spelling, pronunciation, hyphenation, capitalization, usage notes for semantic domains, geographic regions, and propriety; etymological, syntactic and semantic information about the most basic units of the language; example sentences that accompany definitions which often use words in prototypical contexts [41]. In addition, the words are described in terms of senses (lexical concepts), and the concepts are described in terms of words [20].

### D. Information Extraction

The broad definition of information extraction is any method that filters information from large volumes of text. Filtering task retrieves documents from collections and tags particular terms in text [42]. On the other hand, the narrow definition of information extraction is identification and extraction of particular relationships or events in a natural language text. Information extraction creates a structured representation (such as database) of extracted information from the text [42].

Extracted information may be atomic entities and flat records such as entities, relationships, and adjectives or may be richer structures such as tables, lists, and trees from various types of documents [43]. Two main approaches are used in designing information extraction systems: the knowledge engineering approach and the Automatic Training Approach [43]. Other names for the two approaches are hand-coded (or hand-crafted) approach and learning-based approach.

In hand-coded systems, human experts in programming, linguistics, and domain define and develop rules or regular expressions or program snippets for performing the extraction. However, in learning-based system, human experts in a domain manually identify and label representative unstructured examples to train machine-learning models that extract information [44]. An ideal rule-based system has two components [44]: a set of rules and a set of policies that control the firings of multiple rules. The general form of a rule that contain core features of many representative formats.

Rule: Contextual Pattern → Action.

The right-hand side of a rule consists of one or more patterns that have labels and capture various properties of one or more entities and their context in the text. A labeled pattern consists of a pattern and an optional label. A pattern is mostly a regular expression that is defined over features of tokens in the text. The features can be simply any property of the token or its context or its document. The action part of a rule contains various kinds of tagging to assign an entity label to a sequence of tokens, to insert the start or the end of an entity tag at a position, or to assign multiple entity tags [44].

### *E. Sublanguage Analysis*

Barnbrook [10] claimed that the definitions of a dictionary are styled in a distinguished sublanguage that has its own local grammar. Extracting valuable linguistic information from definitions relies on constructing grammars and parsing algorithms for that sublanguage. A sublanguage is a particular natural language type that used in a subject matter or particular domain. A sublanguage has specific features such as specialized vocabulary, semantic relationships, and specialized syntax. Sublanguage examples [45] include weather reports, aircraft repair manuals, scientific articles about pharmacology, hospital radiology reports, and real estate advertisements.

## V. Extracting Information Method

### *A. Experimental Setup*

We selected the Ayn "ع" letter from the Al-Mawrid for both inducing extracting rules and evaluating. The Ayn letter represents about 4.6 % of the Al-Mawrid data volume. The only criteria for selecting the Ayn letter are that its volume is suitable for both the evaluation experiment and the copyright considerations. We used the first half of the Ayn letter to induce the lexical patterns that represent linguistic information to be extracted. The other half is used to evaluate our extraction method.

Two experiments are conducted: the first is to extract both explicit and implicit information in the definitions and the second is to extract the synonymous words and phrases that are used as defining device. In the first experiment, we used the rule-based hand-crafted (or hand-coded) method, while in the other we used simple string processing exploiting some heuristics. Fig. 2 illustrates the phases and steps of the extracting method.

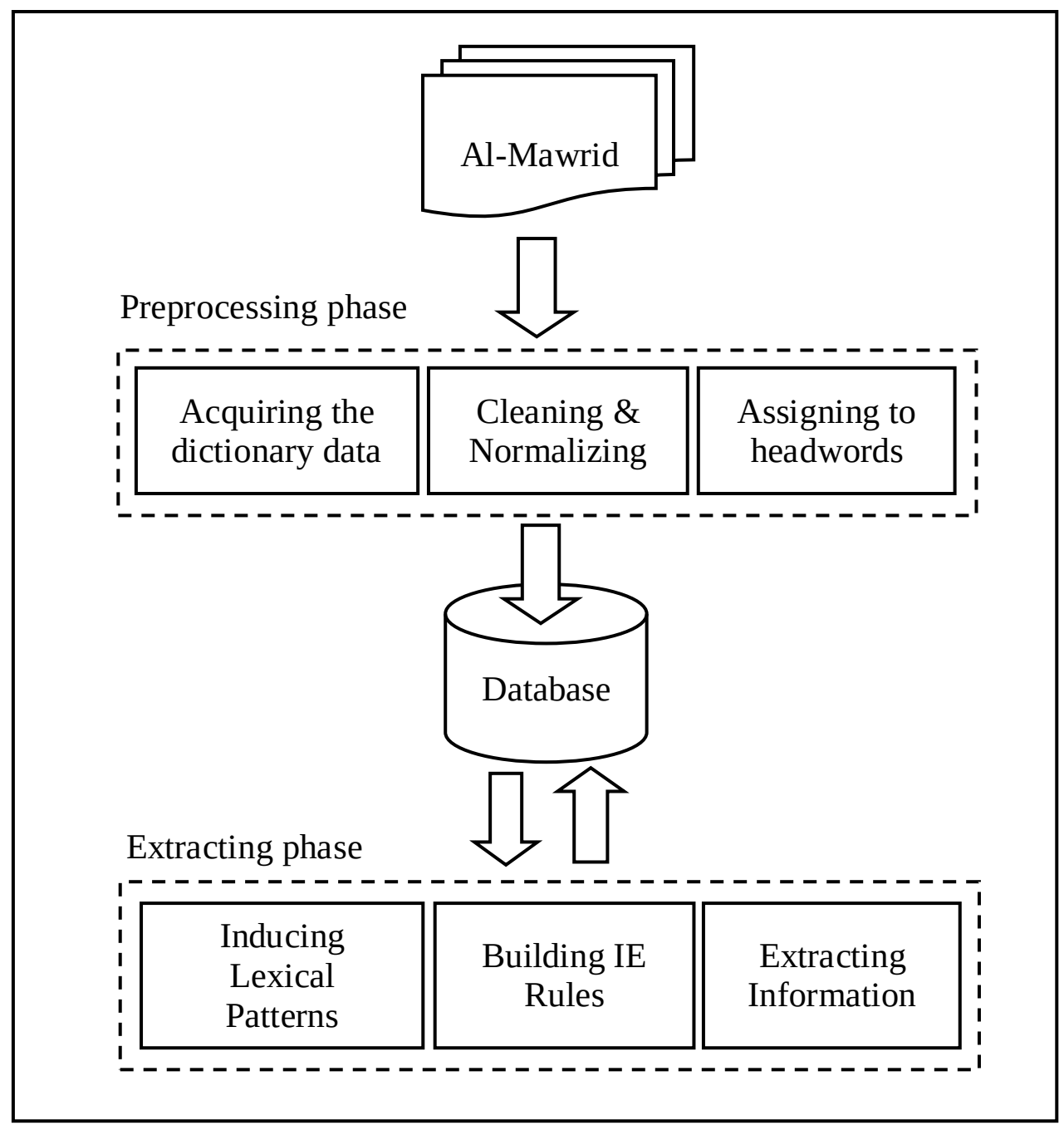

Fig. 2. Phases and steps of the extracting method.

### *B. Experiment I: Extracting of linguistic information*

We subdivided the work in two phases: the preprocessing phase and the extraction phase. The preprocessing phase has three tasks or steps: (a) acquiring the Al-Mawrid data, (b) cleaning & normalizing, and (c) assigning the headwords to their description data (articles). The extracting phase comprises three tasks: (a) inducing lexical patterns and rules, (b) building IE rules, and (c) extracting information. Fiure 2 illustrates the two phases and their relevant steps. We used the same prophase steps that Diaa used in [31].

#### *1) Acquiring the Al-Mawrid Data*

A human data entry explored the software of the Al-Mawrid and copied the entries information of each headword into a text file or spreadsheet file. Each subentry of a headword occupies two consecutive lines in the spreadsheet file, one for Arabic section and the other for the English translation. If English translation is absent, its line is empty.

#### *2) Assigning Headwords to their Articles*

The purpose of this step is to recover the principal database structure of the Al-Mawrid. The principal structure of a dictionary is composed of an index of the headwords and the subentries information that is related to these headwords. Recovering the principal database structure enables us to know the information associated with each headword separately. Consequently, we can make more exploring and processing on the associated information, collecting basic information about the dictionary entries, and assigning the results to the respective headwords. We developed procedures to transform the text files into a rudimentary database of two flat text files: index and data files. See Table 1.

TABLE I. SAMPLE OF THE INDEX AND DATA FILES

| index | |
|---|---|
| ***headword*** | ***#*** |
| عُوَاء | 2568 |
| عَوَائِد | 2569 |
| عَوَائِق | 2570 |
| عَوَادٍ | 2571 |
| عَوَّاد | 2572 |
| عَوَار | 2573 |
| عُوَار | 2574 |
| | 2575 |
| عُوَّار | 2576 |

| data | | | |
|---|---|---|---|
| ***#*** | ***form*** | ***Arabic*** | ***English*** |
| 2568 | عُوَاء | عُوَاء (الكَلْبِ إلخ) | howl(ing),yelp(ing), yowl(ing) |
| 2569 | عَوَائِد | عَوَائِد (مفردها عائِدَة) - راجع عائِدات | |
| 2570 | عَوَائِق | عَوَائِق (مفردها عائِق) - راجع عائِق | |
| 2571 | عَوَادٍ | عَوَادٍ (العَوَادِي) (مفردها عادِيَة) - راجع عادِيَة | |
| 2572 | عَوَّاد | عَوَّاد: عازِفُ العُوْد | lutist, lutanist |
| 2573 | عَوَار، عُوَار | عَوَار، عُوَار: عَيْب، عِلَّة | defect, fault, blemish |
| 2574 | عَوَار، عُوَار | عَوَار، عُوَار: عَيْب، عِلَّة | defect, fault, blemish |
| 2575 | | عُوَار [طب] | anaphylaxis |
| 2576 | عُوَّار | عُوَّار (طائر) | swift, swallow, martin |

*3) Cleaning and Normalizing*

In this step, we carried out three operations on the data: (a) cleaning the data from some faulted characters and un-useful words and, (b) flattening or formalizing some parentheses information, and (c) converting domain labels in the parentheses to labels in square brackets. Examples of un-useful words are "...", "etc.", and "الخ". The Al-Mawrid uses parentheses abundantly and for diverse purposes. The volume and diversity of the parentheses information in the Al-Mawrid is one of many challenges to structure and computerize its data.

Arabic parentheses contain various and multiple information, so they need more deep analysis before we can structure them. This study handled only parentheses that contain labels of semantic or subject domain like "عائِق (نبات)" that is transformed into " عائِق[نبات]". The purpose of this transformation is to normalize all subject labels to be in square brackets, so they can be captured by one IE pattern rule.

*4) Inducing Lexical Patterns*

Based in the sublanguage theory, the defining phrases in the Al-Mawrid are considered a sublanguage of the general language. The defining phrases are produced by one person (the author) or group of persons. Also the defining phrases have special purpose phrases to define headwords of the dictionary. Consequently, the probability of *keywords* or *key phrases* (*formula*) to recurrent frequently for the same purpose is high. Some key words or key phrases are used to manifest morphologic, syntactic information, or semantic relations.

We used word frequency distribution and keyword in context (KWIC) analysis as the main analysis devices to discover patterns in the training defining phrases corpus. Then, the discovered patterns are used to design information extraction rules that are used to extract information in other defining phrases corpus.

We built specialized tool to facilitate this analysis. contains some result samples of the analysis. The fourth column contains knowledge that we induced after analysis. For example, in the pattern "noun: "مصدر" + past verb", the keyword "مصدر" is used in the definitions of a noun headword. It also always followed by past verb. The key word "مصدر" manifests morphological relation between a noun headword and past verb; the morphological relation is derivation relation (علاقة اشتقاق) in the Arabic language. Table 2 shows samples of the Analysis.

After analyzing the half of Ayn "ع" letter of the Al-Mawrid, we discovered some shortcomings in this dictionary. Generally, Arabic dictionaries have problems with some aspects that lexicographers should solve [46]. These problems are challenges for both extracting information and computerizing Arabic dictionaries. Deficiencies exits in the Arabic dictionaries at three levels [46]: the theoretical foundation, the analysis, and the representation of syntactic-semantic content. In addition, The Al-Mawrid has some special deficiencies. The following are some shortcomings in the Al-Mawrid that complicate extracting information task.

*Inconsistency*:

- Domain labels exist either in bracket parentheses [نبات] or in parentheses (نبات).
- A semantic relation may have more than one key word that manifests it. For example, the antonym relation is expressed by any of the words "ضد", "عكس", and "لا".
- The adjective "صفة" tag and name "اسم" tag are the only part-of-speech (POS) tags in the Al-Mawrid. The percentage of POS tags approach to zero.
- There is no a consistent clue that determines whether a subentry is either a meaning of the headword or an example, a collocation, an idiom, etc. A colon mark distinguishes the meaning only if it exists after a copy of the headword.
- The explanation in a subentry may be expressed by phrases that follow a colon mark, by phrases in parenthesis like "عقيد (في الجيش)", or by nothing at all.
- That information does not exist in the definitions in consistent way. Not all the subentries contain the same types of information. In other words, there is no a fixed set of linguistic information that exist in each subentry. All explanation fields or header fields approximately contain no or one type of information that can be extracted.

*Cross-reference ambiguity*: A cross-reference word or phrase may refer to more than one headword and each headword may have more than one subentry. Furthermore, a subentry may represent either a meaning of the headword or an example, an idiom, a term, a restricted collocation, etc.

*Parentheses*: The parentheses information has large varieties and has multiple purposes of functions. For example, the parentheses may contain prepositions, synonyms, examples, explanations, antonyms, morphological information, spelling variations, etc.

*Indefinite keywords:* The author of the Al-Mawrid does not cite comprehensive list of keywords that manifest morphologic, syntactic, or semantic information in the introduction of Al-Mawrid, so we need to discover these keywords before either structuring definitions or extracting information.

*Randomly scattered information:* The syntactic and semantic information in the defining phrases are scattered randomly in the defining phrases.

*5) Building Extracting Rules*

Assuming the definitions of the Al-Mawrid follow sublanguage characteristics, we can use the rule-based hand-crafted information extraction with patterns described as word sequences directly. This what Grishman stated in [47] “Because of the complexity of natural language (except in the most restricted of sublanguages), it is not practical to describe these patterns directly as word sequences.”

Since the sample of data is small, we found the rule-based information extraction is more suitable to extract the lexical patterns. We developed the rules based on the lexical patterns that are discovered in the previous sections. The IE rule is composed of two parts: the contextual pattern and the action. The contextual pattern is implemented by regular expression. The action part comprises two sub-actions: the matching action and the filling action. The matching action matches the trigger string of the contextual pattern with matcher string in the template. The filling action fills the extracted strings from the contextual pattern into output record. Figure 4 illustrates the structure of a template. A template consists of matching, information type, POS, relation type and its sites, and two filling slots. The first four slots are prefilled according to analysis steps. The matching slot contains a string (English name) that matched with a trigger string in the regular expression. The two filling slots are filled with extracted strings in the regular expression. See Fig. 3 for examples of extracting information rules.

*6) Extracting Information*

The process of extraction is performed in two steps by two components: the REs component and the templates component. The REs component is a set of regular expressions. The templates component is a set of templates that contains slots. The information that already prefilled in the templates is based on a phenomenon in the Al-Mawrid definitions; some *key words* or *recurrent formula* are occurred only in specific part of speech tags and manifest some specific morphologic, syntactic, or semantic information. The relation between that key words and that knowledge is one-to-one relation. We can formulate that relation as function. Based on that phenomenon, we designed a set of templates. Each template composes sex slots. The first four slots are prefilled with information that is particular to a specific *key word* or *formula*. The other two slots are filled by the extracted information from an Arabic definition. See Fig. 4.

We considered the keywords that manifest relations or provide linguistic information as entities. Some entities represent an independent knowledge like domain “[كيمياء]” and others represent relations such as morphologic, syntactic, or semantic relations. Most of the relations are binary relations. So we set up the problem of extracting lexical information from the Al-Mawrid as information extraction problem where the entities and the relations are keywords in the defining phrases. The positional relations between the keywords and the other words in the definitions that compose lexical patterns take two forms: binary relation and unary relation. In the binary relation, there exist a left filler and right filler in the lexical pattern. On the other hand, the unary relation has only one filler that may be the right or the left filler. Fig 4 shows an extracting step for a definition.

TABLE II. EXAMPLES OF ANALYSIS STEP

<table>
<tr><th>1-gram word pattern</th><th>#</th><th>Definition Examples & Patterns & induced knowledge</th></tr>
<tr><td>مَنْ</td><td>12</td><td>مُؤَجِّر: مَنْ يُؤَجِّر | مُدَلِّكَة: مَنْ تُدَلِّك<br>noun: "مَنْ" + present verb<br>Derivation Relation (إشتقاق): ( يُؤَجِّر , مُؤَجِّر)<br>Attribute: Human</td></tr>
<tr><td>مَصْدَر</td><td>7</td><td>تَأبِير: تَلْقِيح، مَصْدَر أَبَّرَ | اِبْتِغاء: مَصْدَر اِبْتَغَى<br>noun: "مصدر" + past verb<br>Derivation Relation (إشتقاق): ( أَبَّرَ ,تَأْبِير)</td></tr>
<tr><td>[طب]</td><td>6</td><td>عَصَّبَ: زَوَّدَ بِأعْصاب<br>past verb: “زَوَّدَ بِ” + noun<br>Derivation Relation: (أَعْصاب , عَصَّبَ)<br>Domain: medicine</td></tr>
<tr><td>ج</td><td>5</td><td>مُقْتَنًى (ج مُقْتَنَيات): ما اقْتُنِيَ<br>singular (ج plural)<br>Plural Relation: ( مُقْتَنَيات , مُقْتَنًى)</td></tr>
<tr><td>ضِدّ</td><td>4</td><td>اِنْقِطَاع: ضِدّ اتِّصال | أسْوَد: ضِدّ أبْيَض<br>adj.: "ضِدّ" + adj. | noun: "ضِدّ" + noun<br>Antonym Relation: (أَبْيَض ,أَسْوَد)</td></tr>
</table>

*C. Experment II: Extraction of Synonyms*

The author of the Al-Mawrid used synonyms as main device to define headwords. The Author used punctuations for manifesting implicitly synonymous words or phrases. Arabic comma is used to separate synonymous words or phrases. Also the conjunction word “أو” (or) is used sometimes to indicate that more than one or phrases are synonyms.

The following are different forms of synonyms in the definitions. (a) synonyms may be between parentheses such as “دارُ (أو مَأْوَى) العَجَزَة”; (b) synonyms may be coma-separated words or phrases in the explanation section of the definition such as “عَبْقَرِيَّة: نُبُوغ، ذَكَاءٌ عالٍ”; (c) synonyms may be coma-separated words or phrases in the header section of the definition such as “عَبَاءَة، عَبَاء: رِدَاءٌ طَوِيلٌ واسِع”. In this case the synonyms usually are word-form variations for the same word; and (d) the forms of the conjunct or “أو” Note: we consider two criteria to consider a phrase as synonym: the phrase is composed of at maximum two words; one of the two words may be a preposition.

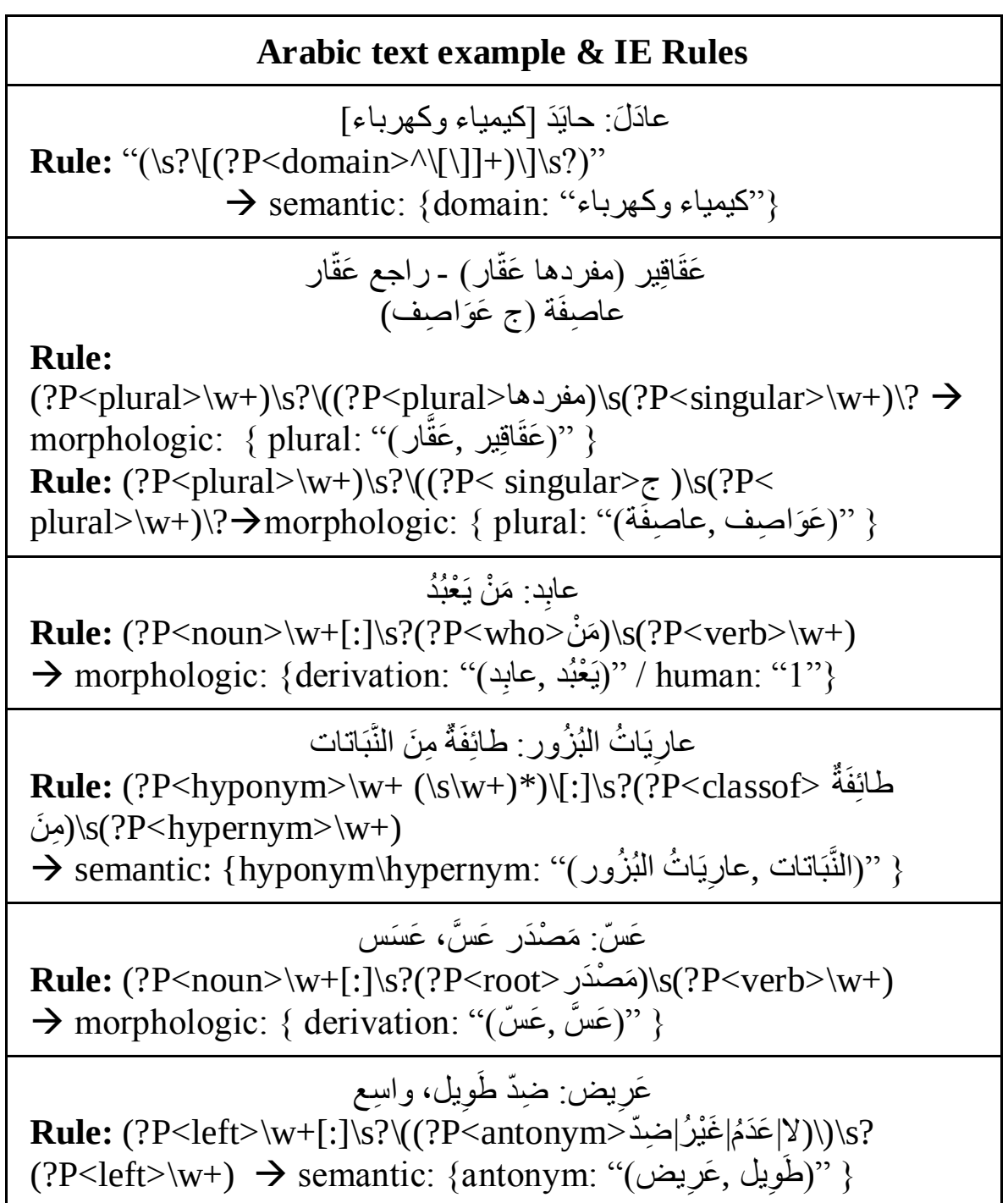

| Arabic text example & IE Rules |
|---|
| عادَل: حايَدَ [كيمياء وكهرباء]<br>**Rule:** "(\s?\[(?P<domain>^\[\]]+)\]\s?)"<br>→ semantic: {domain: "كيمياء وكهرباء"} |
| عَقَاقِير (مفردها عَقّار) - راجع عَقّار<br>عاصِفَة (ج عَوَاصِف)<br>**Rule:**<br>(?P<plural>\w+)\s?\((?P<plural>مفردها)\s(?P<singular>\w+)\)? →<br>morphologic: { plural: "(عَقَاقِير ,عَقّار)" }<br>**Rule:** (?P<plural>\w+)\s?\((?P< singular>ج )\s(?P<<br>plural>\w+)\)?→morphologic: { plural: "(عَوَاصِف ,عاصِفَة)" } |
| عابِد: مَنْ يَعْبُدُ<br>**Rule:** (?P<noun>\w+[:]\s?(?P<who>مَنْ)\s(?P<verb>\w+)<br>→ morphologic: {derivation: "(يَعْبُد ,عابِد)" / human: "1"} |
| عارِيَاتُ البُزُور: طائِفَةٌ مِنَ النَّبَاتات<br>**Rule:** (?P<hyponym>\w+ (\s\w+)*)\[:]\s?(?P<classof> طائِفَةٌ<br>مِنَ)\s(?P<hypernym>\w+)<br>→ semantic: {hyponym\hypernym: "(النَّبَاتات ,عارِيَاتُ البُزُور)" } |
| عَسّ: مَصْدَر عَسَّ، عَسَس<br>**Rule:** (?P<noun>\w+[:]\s?(?P<root>مَصْدَر)\s(?P<verb>\w+)<br>→ morphologic: { derivation: "(عَسَّ ,عَسّ)" } |
| عَرِيض: ضِدّ طَوِيل، واسِع<br>**Rule:** (?P<left>\w+[:]\s?\((?P<antonym>لا\|عَدَمُ\|غَيْرُ\|ضِدّ)\)\s?<br>(?P<left>\w+)  → semantic: {antonym: "(طَوِيل ,عَرِيض)" } |

Fig. 3. Instances of extracting rules.

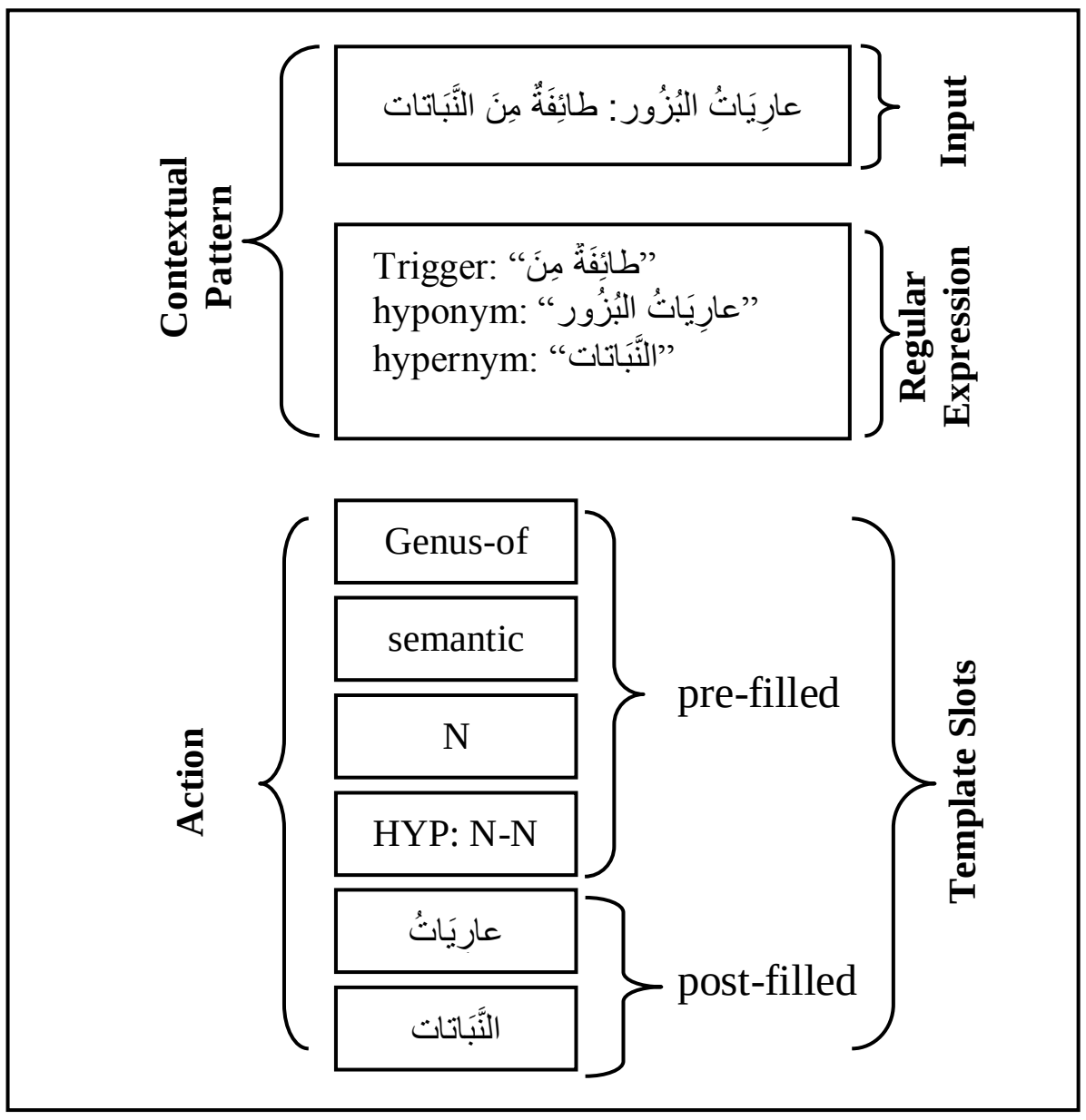

Fig. 4. An example of extracting step.

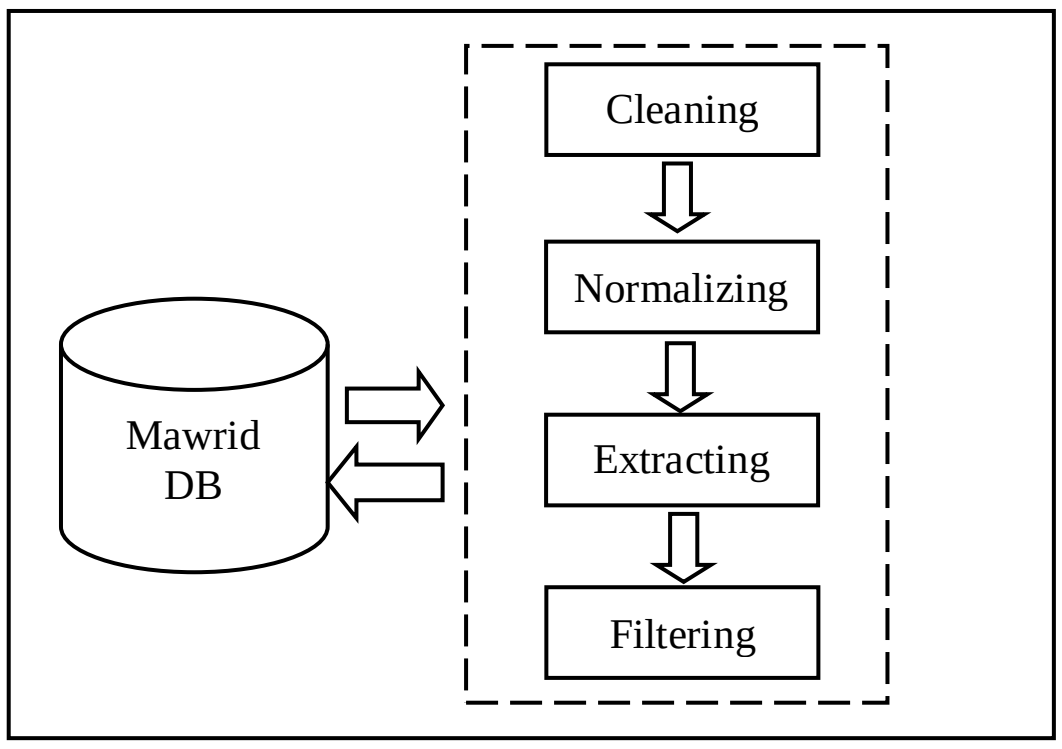

Fig. 5. Module of synonyms extractor.

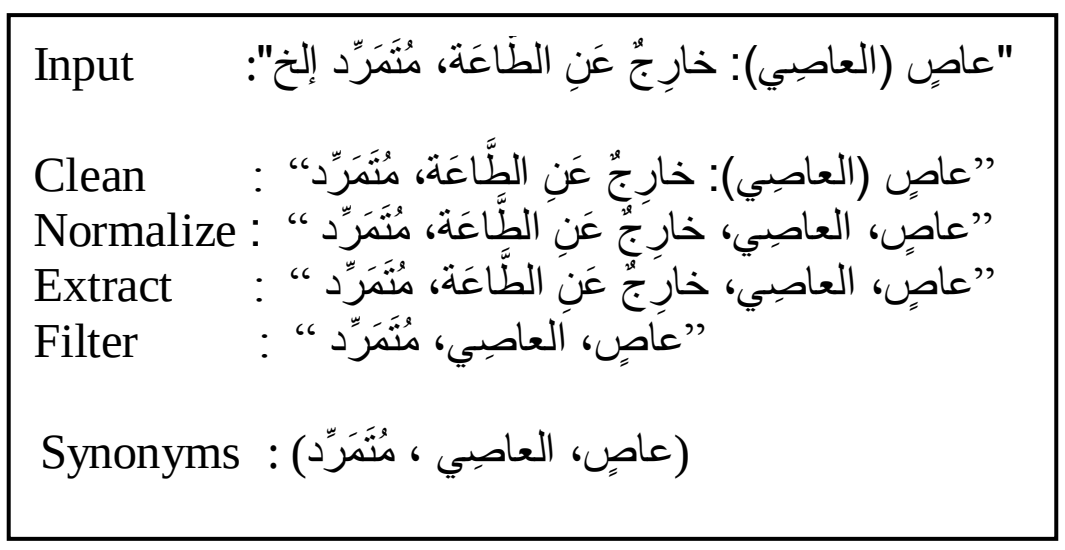

Fig. 6. Synonymous extraction example.

Fig. 5 illustrates the steps of extracting synonyms words or phrases and Fig. 6 shows an extraction example. The cleaning step removes any un-useful words or characters. The normalizing step reduces the variations of synonym punctuations and forms to decrease the number of matching-pattern extractors that are required. The filter step removes the defining phrases or lexical patterns that are previously extracted in the previous module. Also, any defining phrases that have more than three words will be considered explanation not synonyms; so it will remove.

## VI. Results and Discussion

We can compute the gross scoring of an IE system by calculating the weighted average over all results. The most common measures to evaluate an IE system are *recall*, *precision*, and *F-measure*. Manning 1999, calculated recall and precision from the confusion matrix [48].

$$Recall = true\ positive / (true\ positive + false\ negative)$$
$$Precision = true\ positive / (true\ positive + false\ positive)$$

Recall shows the number of items that is extracted effectively and that should have been found. Precision shows the number of correct items that is extracted.

### *A. The synonyms evaluation*

Table 3 contains the evaluation figures of precision and recall for the extracted information. The high precision value indicates that the rules and its regular expression patterns are designed and implemented in high precision value. The high recall value indicates that the synonyms are used extensively in the definitions.

TABLE III. EVALUATION RESULTS

| | Precision | Recall |
|---|---|---|
| **synonyms** | 85% | 98 % |
| **other information** | 88 % | 73% |

## B. *The other information evaluation*

By investigating the cause of low recall value, we found that the hyponym/hypernym relations are two types: closed set that have low numbers of keywords that manifest them and open set that constitutes most definitions of the dictionary. The open set is definition phrases composed of two parts: *genus* and *differentia*. The genera (plural of genus) are open set that we cannot compute and build IE rules.

## C. *The percentage of extracted information*

Fig. 7 shows the relative and absolute percentage of extracted information. The relative percentage is computed according to total number of extracted information from all types. The absolute percentage is computed according to the total number of subentries of the Ayn “عَيْن” letter. The chart reveals that the most significant information in the Al-Mawrid is synonyms, domain, derivation, and hyponym/hypernym.

The hyponym/hypernym relation is the most important one in MRDs [20]. This relation is used to build lexical taxonomies and semantic networks. So, we investigate this relation in more deep. We found that two types of hyponym/hypernym relation: closed (or explicit) and open (or implicit). The explicit relation is represented by key words or phrases that manifest it and they have about 33% relative to the total number of HYP relation. The other type of HYP is open and has no key words to manifest them. They have 77% relative to the HYP relations. Because there are no key words that represents this type the recall percent is low.

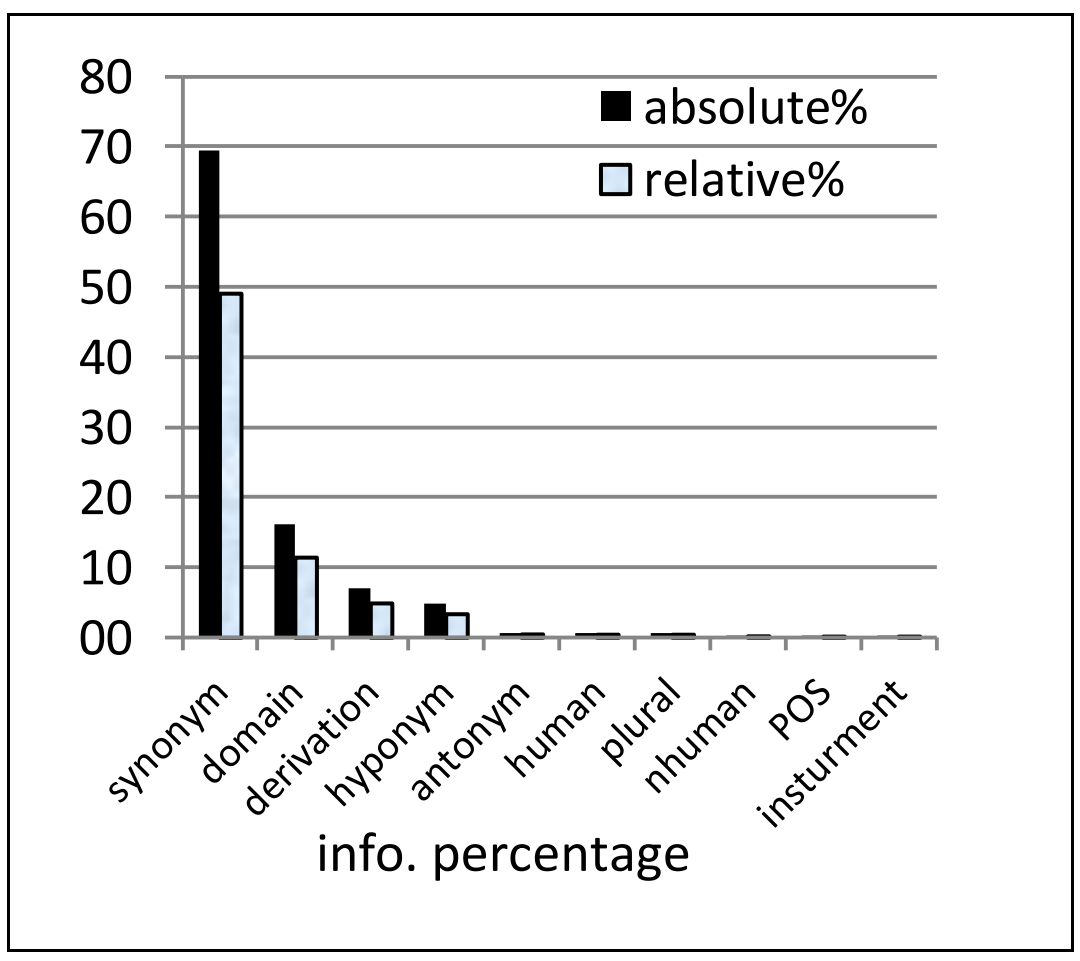

Fig. 7. Absolute and relative numbers of extracted info.

## D. *The Caveat of this Paper*

- The study includes only the Arabic section of the Al-Mawrid definitions.
- The semantic relations between verbs have not been involved.
- The Arabic letter chapter of “ع” has biasing subentries that affect the percentage of domain. There is a lexical pattern “<Noun> + علم” that have many number of subentries. Example of that pattern is the entry “عِلْمُ النَّفْس psychology”. In the analysis, we assigned a domain label to such entries.
- The study is not comprehensive. Other information can be extracted like named entities (NEs), multiword expressions (MWs), etc.

# VII. CONCLUSIONS AND FUTURE WORK

The Arabic section of the Al-Mawrid definitions contains significant amount of synonyms, domains labels, derivations, and hyponym/hypernym relations. The semantic information (synonyms, domains, hyponym/hypernym) is the more significant and important ones in the Al-Mawrid. Most of the information is scattered randomly and inconsistently in the definitions. Some information has both closed set and open set variations; only the closed set that can be extracted, in high precision and recall, by rule-based information extraction. The Al-Mawrid cannot be used independently to build practical computational lexicon for real NLP application. The author of the Al-Mawrid used synonyms as the main device for defining and edxplaining headwords.

In future study, we will try improve recall percentage of the open set (implicit) hyponym/hypernym relations. The parentheses contain rich semantic information that needs to be analyzed and extracted. We will extract other information from the Al-Marid definitions like named entities (NEs), multiword expressions (MWs), etc. Furthermore, the English section of the Al-Mawrid needs to be studied in standalone work to extract information.

TABLE IV. PERVENTAGE OF CLOSED AND OPEN HYP

| Closed set (hyponym/hyernym) HYP | Open set (hyponym/hyernym) HYP |
|---|---|
| 33% | 77% |